\documentclass{article}
\usepackage{ijcai24}

\usepackage{times}
\usepackage{soul}
\usepackage{url}
\usepackage[hidelinks]{hyperref}
\usepackage[utf8]{inputenc}
\usepackage[small]{caption}
\usepackage{graphicx}
\usepackage{amsmath}
\usepackage{amsthm}
\usepackage{booktabs}
\usepackage{algorithm}
\usepackage{algorithmic}
\usepackage[switch]{lineno}

\usepackage{algorithm}
\usepackage{algorithmic}
\usepackage{graphicx}
\usepackage{amsmath}
\usepackage{amssymb}
\usepackage{booktabs}
\usepackage{subcaption}
\usepackage{multirow}
\usepackage{stfloats}

\title{BlockEcho: Retaining Long-Range Dependencies for Imputing Block-Wise Missing Data}

\author{
Qiao Han
\and
Mingqian Li\and
Yao Yang\And
Yiteng Zhai$^{*}$
\\
\affiliations
Zhejiang Lab\\
\emails
\{hanq, mingqian.li, yangyao, ito\}@zhejianglab.com,
}

\begin{document}

\maketitle

\begin{abstract}
    Block-wise missing data poses significant challenges in real-world data imputation tasks. Compared to scattered missing data, block-wise gaps exacerbate adverse effects on subsequent analytic and machine learning tasks, as the lack of local neighboring elements significantly reduces the interpolation capability and predictive power. However, this issue has not received adequate attention. Most SOTA matrix completion methods appeared less effective, primarily due to overreliance on neighboring elements for predictions. We systematically analyze the issue and propose a novel matrix completion method ``BlockEcho" for a more comprehensive solution. This method creatively integrates Matrix Factorization (MF) within Generative Adversarial Networks (GAN) to explicitly retain long-distance inter-element relationships in the original matrix. Besides, we incorporate an additional discriminator for GAN, comparing the generator's intermediate progress with pre-trained MF results to constrain high-order feature distributions. Subsequently, we evaluate BlockEcho on public datasets across three domains. Results demonstrate superior performance over both traditional and SOTA methods when imputing block-wise missing data, especially at higher missing rates. The advantage also holds for scattered missing data at high missing rates. We also contribute on the analyses in providing theoretical justification on the optimality and convergence of fusing MF and GAN for missing block data.

\end{abstract}

\section{Introduction} \label{introduction}

The issue of missing data is prevalent in real-world datasets, stemming from factors such as users' reluctance to share data, variations in feature structures, privacy concerns, and unintended data corruption over time \cite{scheffer2002dealing}. In scenarios involving large-scale and highly correlated missing data, matrix completion becomes essential to impute the unknown entries accurately and efficiently. Accurately and efficiently estimating the unknown elements in a matrix is pivotal for effective dataset analysis and supports subsequent machine learning and operations research tasks. Matrix completion finds wide application in recommender systems, trajectory recovery, phase retrieval, computer vision, genotype imputation and other fields of research \cite{santos2019generating}. Missing data isn't limited to randomly scattered elements; it can also occur in blocks, referred to as block-wise missing data. For example, discrepancies in feature sets provided by different agents can result in non-overlapping blocks of NA values during data aggregation. Moreover, certain regions may experience skipped clinical trials due to technical and financial constraints, while continuous data corruption can arise from malfunctioning monitoring devices over time.

A diverse array of methods have been developed over the years to address the matrix completion problem, spanning traditional techniques like linear interpolation, K-nearest neighbors imputation, Multiple Imputation by Chained Equations (MICE), and ensemble methods such as MissForest \cite{williams2015missing}. In the new data regime, Generative Adversarial Networks (GANs) have also demonstrated immense potential in accurately estimating high-dimensional data distributions \cite{goodfellow2020generative}. Consequently, recent works have explored GAN architectures for data recovery in the missing data context. However, most prevailing completion techniques display suboptimal effectiveness on block-missing matrices, especially with increasing missing proportions. They predominantly leverage neighboring available elements to predict unknown entries, making them vulnerable to systematic feature gaps. An exception is Matrix Factorization (MF) which implicitly retains inter-sample dependencies across longer ranges \cite{hastie2015matrix}. But its linearity assumptions limit capturing complex nonlinear relationships. To unite their complementary strengths and offset limitations, we propose \textbf{BlockEcho} - an integrated approach tailored for block-missing data that capitalizes on the strengths of both GAN and MF. It explicitly retains long-range inter-element associations via MF while modeling intrinsic nonlinear patterns through GANs. The key contributions are:


\begin{enumerate}
\item  We have conducted an in-depth analysis and formally defined the concept of ``Block-wise" missing data. Subsequently, we have innovatively introduced a solution to address this challenge, named BlockEcho.
\item We extensively benchmark BlockEcho against SOTA baselines on diverse public datasets for traffic, epidemiology, and recommendations. Results demonstrate significant gains in imputation accuracy under both block-missing and high-rate scattered missing regimes. Downstream forecasting tasks further showcase advantages.
\item We provide theoretical analysis into global optimality and convergence for the proposed integrated objective. We particularly justify the synergistic effects of MF encoding long-range dependencies and GANs locally adapting complex distributions. Rigorous proofs supplement empirical observations.
\end{enumerate}

The rest of this paper is structured as follows. In Section \ref{literature_review} we provide a concise literature review in missing data imputation studies. Section \ref{pro} defines the problem. We elaborate our model BlockEcho in Section \ref{methodology}. Section \ref{theory} provides a theoretical discussion. Section \ref{experiment} details our experiments and  results. We conclude the paper in Section \ref{conclusion}.

 \begin{figure*}[t]
\centering
\includegraphics[width=0.9\textwidth]{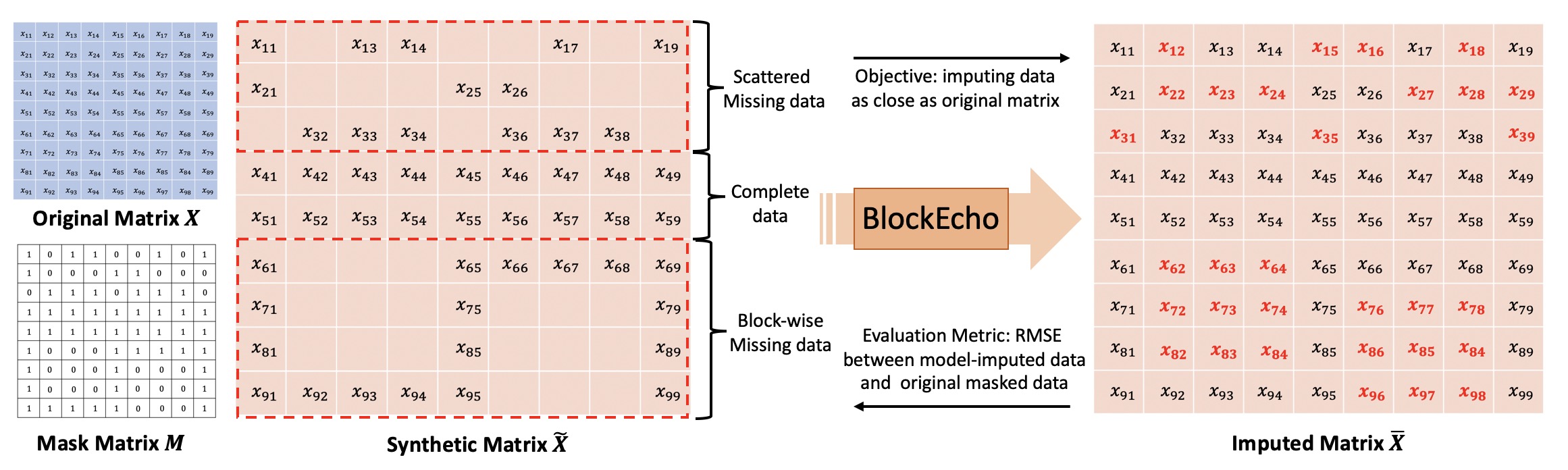} 
\caption{Problem definition, obejective and evaluation metric}
\label{fig:matrix}
\end{figure*}

\section{Related Work} \label{literature_review}

\subsection{Data imputation approaches}
Broadly, prevailing techniques for missing data imputation fall under two categories – discriminative or generative. Discriminative methods directly estimate the conditional distribution of missing values given observed entries. This includes MICE \cite{2000Multivariate}, MissForest \cite{2011MissForest} and Matrix Factorization \cite{hastie2015matrix} \cite{yu2016temporal}. Generative approaches instead model the complete joint data distribution and draw imputes from conditionals on observed parts. Traditionally, expectation maximization algorithms iteratively refining estimates were common \cite{gold2000treatments}. Recently, advances in deep generative modeling have spawned powerful imputation paradigms. Notably, J. Yoon et al. \cite{yoon2018gain} pioneered Generative Adversarial Networks (GANs) for missing data imputation through the GAIN framework in 2018. Instead of assuming explicit parametric forms, GANs can flexibly approximate arbitrary data distributions. Consequently, GAN-based solutions have risen as SOTA, significantly improving over pre-GAN techniques.

\subsection{GAN-based methods}
Many works have since built upon GAIN, either enhancing its convergence like SGAIN (slim Wasserstein GAIN) \cite{neves2021sgain}, integrating auxiliary models as in GAMIN (Generative adversarial multiple imputation network) \cite{yoon2020gamin}, or exploiting complementary data modalities like images via CollaGAN (Collaborative GAN for missing image data imputation) \cite{lee2019collagan}. Some explore sophisticated relationships within data beyond distributions, including: 1) Dual generative modeling of values and high-dimensional feature abstractions in MisGAN \cite{li2019misgan}, 2) time-series self-attention mechanisms in \cite{zhang2021missing} to capture sensor inter-dependencies, 3) Pretraining classifier-guided generative models as PC-GAIN \cite{wang2021pc}, 4) Distribution-centered losses like STGAN \cite{yuan2022stgan}, and 5) Multimodal spatio-temporal modeling via GAN+RNN+GCN \cite{kong2023dynamic}. ANODE-GAN \cite{chang2023anode} also augments GANs with auxiliary variational autoencoders.

Nonetheless, prevailing GAN imputation approaches overlook systematic data deficiencies like block-missing data. Mostly relying on local neighborhoods, they remain less effective as gaps exacerbate, especially at higher missing rates. Our proposed BlockEcho framework aims to address this limitation by uniquely blending GANs with matrix factorization.

 \section{Problem Definition} \label{pro}
 


Consider an original data matrix $X \in \mathbb{R}^{m*n}$. $M \in \{0,1\}^{m*n}$ is set as a binary mask matrix codifying missingness, where $M_{ij}=0$ denotes element $X_{ij}$ as unobserved. Imposing $M$ on $X$ gives the incomplete matrix:

\begin{equation}
\label{eq0} 
\tilde{X}_{ij} =\begin{cases}
X_{ij} & if M_{ij}=1
\\NaN & if M_{ij}=0\\
\end{cases}
\end{equation} 


We refer to $X$ as the original matrix, $M$ as the mask matrix, and $\tilde{X}$ as the missing data matrix. The data imputation problem aims to obtain an estimation $\hat{X}$ of $X$ given $\tilde{X}$ and $M$, with the objective of minimizing the discrepancy between the imputed data and the original data. Our final output is the imputed matrix $ \bar{X} $ defined as
\begin{equation}
\label{mix_row} 
\bar{X}:=\tilde{X}\odot M + \hat{X} \odot (1-M) 
\end{equation} 
where $\odot$ denotes element-wise multiplication. 

For any given coordinates $(i_{l},j_{l})$, $\exists \  i_{u} \geq i_{l}+3$ and $j_{u} \geq j_{l}+3$, $\forall \   i_{l} \leq i \leq i_{u}$ and $j_{l} \leq j \leq j_{u}$, $M_{ij}=0$, then we define this area as exhibiting a local block-wise missing pattern. Following the definition, the missing data in datasets can manifest as uni-block-wise, multiple-separate-block-wise, scattered, or as a mixed combination thereof.

We conducted a straightforward experiment to understand the harm of missing data.  We employed a random forest algorithm to forecast the next timestamp's traffic flow for dataset PE-BAY (outlined in Section \ref{conclusion}), without any feature engineering but normalization. Each category of missing data exhibited a 60\% missing rate. Our experimental findings, as presented in Table \ref{result_wh}, highlighted the notably adverse impact of a substantial, uniformly absent data block on the accuracy of predictions. Henceforth, unless explicitly specified, ``block-wise missing" denotes the uni-block.

  

\begin{table}[htbp]
    \centering  
    \begin{tabular}{rrrr}
    \toprule
    Ori-Data&Scattered&Tri-Blocks&Uni-Block\\ 
    \midrule    
     0.0395& 0.0961& 0.1132& 0.1249\\
    \bottomrule
  \end{tabular}
  
  \caption{WMAPE of prediction task in different missing type 
  }
  \label{result_wh}
\end{table}

\begin{figure*}[t]
\centering
\includegraphics[width=0.85\textwidth]{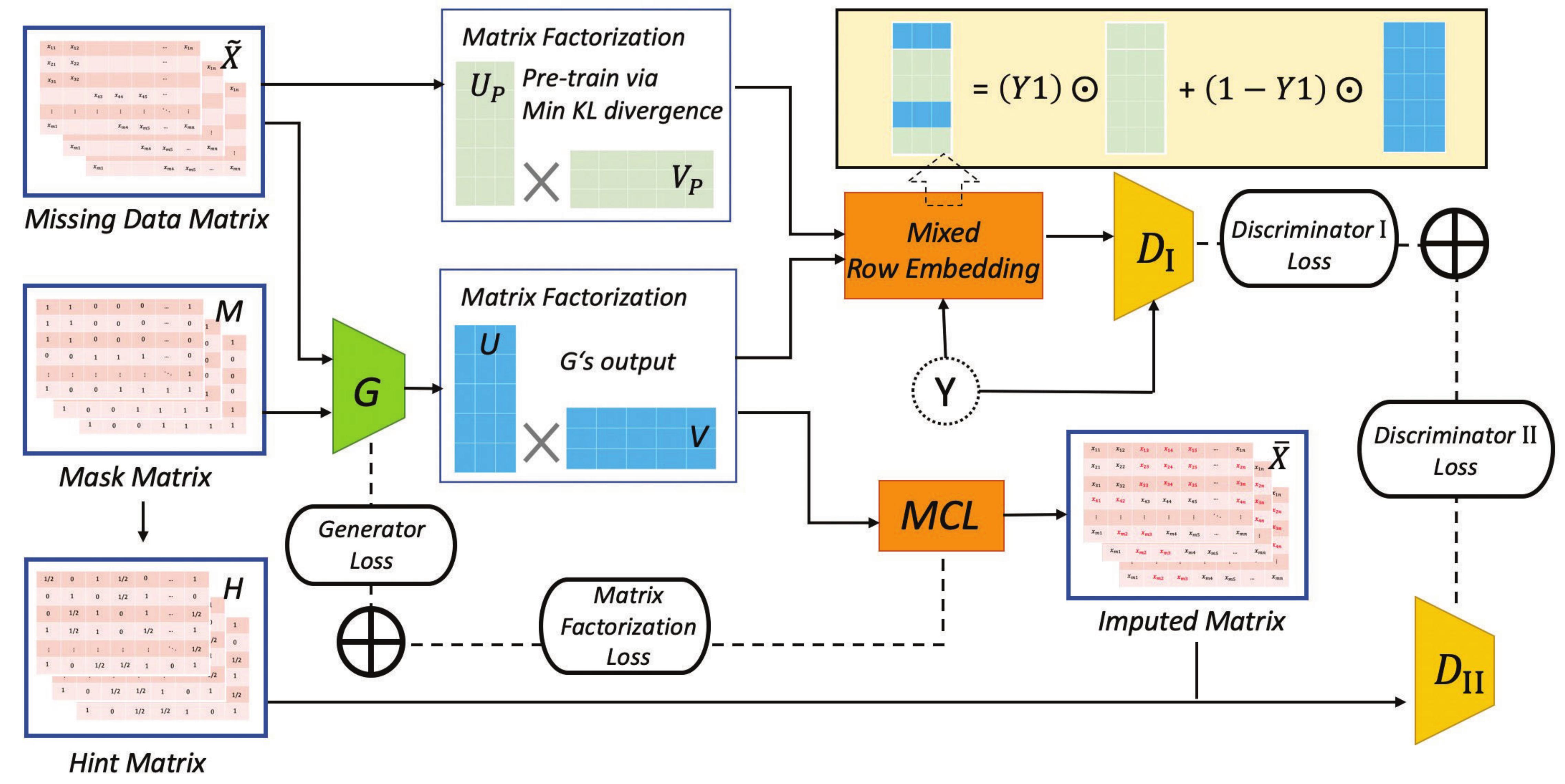} 
\caption{The architecture of BlockEcho}
\label{fig:architecture}
\end{figure*}


\section{The Architecture of ``BlockEcho"} \label{methodology}
In this section, we will elaborate on our integrated model, elucidating how the MF and GAN components synergize to enhance overall performance. The overarching model architecture of BlockEcho is depicted in Figure \ref{fig:architecture}.

\subsection{Matrix Factorization (MF) and Pre-training}  \label{mfa}
Within the MF component, the missing data matrix is decomposed into two smaller matrices, denoted as the imputed row embedding matrix $U$ and the imputed column embedding matrix $V$. Their dimensions are $m*h$ 
(adjustable to mini\_batch\_size$*h$ during training) 
and $h*n$, where $h$ is a hyperparameter. In essence, $h$ serves as an indicator of the original matrix's level of freedom, with smaller values reflecting a higher degree of interdependency among the data in the original matrix.

As illustrated in Figure \ref{fig:architecture}, there are two MF components in BlockEcho. One is deployed during the pre-training stage, breaking down the matrix into $U_p$ and $V_p$ to extract long-range relationship features, which are then used to supervise model training in a transformed space through a discriminator $D_I$.  The other is directly involved in neural network training and result generation: the imputed row embedding matrix $U$ is produced by the generator, while the imputed column embedding matrix $V$ is directly calculated through backpropagation. Subsequently, these matrices undergo a matrix completion layer (MCL) to match the known parts of the missing data matrix. Rather than employing direct multiplication, MCL incorporates a fully connected neural network layer after multiplication. . This technique is employed to introduce nonlinear relationships while upholding the inherent advantage of low constraints on the degree of freedom in the matrix factorization method. Both MF components are oriented towards minimizing the Kullback-Leibler divergence of known elements, thereby establishing the MF objective as
 \begin{equation}
 \label{loss1}
\begin{split}
\min \ \  &\sum_{i,j}((X \odot M)_{ij}log(\frac{(X \odot M)_{ij}}{(MCL(UV) \odot M)_{ij}})
\\&-(X \odot M)_{ij}+(MCL(UV) \odot M)_{ij})
\end{split}
\end{equation} 

\subsection{Generative Adversarial Nets (GANs)} \label{gan}
The GAN framework consists of a generator $G$ and two discriminators $D_I, D_{II}$ as denoted in Figure \ref{fig:architecture}. The generator $G$ takes $\tilde{X}$ with 0 imputed, $M$, and a noise matrix $Z$ with the same size as $U$ as the inputs, and $U$ as the output. Next, MCL takes $U$ as input and outputs the estimation $\hat{X}$. With $\hat{X}$, $\tilde{X}$ and $M$, we build the imputed matrix $ \bar{X} $ following Equation (\ref{mix_row}). Given a vector of length h, the discriminator $D_{I}$ will determine whether it originates from pre-trained matrix factorization or from the generator. To do this, we randomly generate $Y\in \{0,1\}^{m*1}$ as the index vector and build a mixed-row matrix $U_D$ combining $U_p$ and $U$: 
\begin{equation}
U_D = (YI) \odot U_p+(1-YI) \odot U
\end{equation} 
where $I\in \{1\}^{1*h}$. The discriminator $D_{I}$  will output a result of the same size as $Y$ and the objective is
\begin{equation}
\label{loss2}
\mathop{\min}\limits_{G} \mathop{\max}\limits_{D_{I}}
\ Y^T logD_I(U_D)+ (1-Y)^T log(1-D_I(U_D)) 
\end{equation} 

The discriminator $D_{II}$  is designed to discern whether the elements in $ \bar{X} $ is from $\tilde{X}$ (real) or $\hat{X}$ (fake). This differs from a standard GAN model which aims to differentiate inputs ranging from entirely true to entirely false. In addition to $ \bar{X} $, $D_{II}$ incorporates the hint matrix $H$ as input. Hint matrix, proposed in GAIN method \cite{yoon2018gain}, modifies the elements of $M$ to 1/2 at a specified rate. This modification facilitates $D_{II}$ to converge more swiftly and accurately by furnishing ``enough" information regarding $M$. 
The objective of this component is 
\begin{equation}
\label{loss3}
\begin{split}
\mathop{\min}\limits_{G} \mathop{\max}\limits_{D_{II}} &
\ \sum_{i,j}(M \odot logD_{II}(\bar{X} , H)
\\&+ (1-M) \odot log(1-D_{II}(\bar{X}, H)))_{ij} 
\end{split}
\end{equation} 

In particular, the MF component serves as a structured constraint and guidance for the GAN component (analogous to the relationship between CNN and MLP), enabling the entire network to explicitly preserve long-range relationships (similar to CNN's explicit preservation of proximity relationships). This fundamental characteristic underpins the superior performance of our model compared to baselines on datasets with missing blocks or high missing rates.

\subsection{Objective}
According to (\ref{loss1}), (\ref{loss2}) and (\ref{loss3}), the objective of the whole multi-loss model is 


\begin{equation}
\label{eq5}
\resizebox{.98\linewidth}{!}{$
\begin{split}
  \mathop{\min}\limits_{G} \mathop{\max}\limits_{D}  &
 \ \  (1-\alpha) [(1-Y)^T log(1-D_I(U_D))
  \\& + Y^T logD_I(U_D)+  \sum_{i,j}
 (M \odot logD_{II}(\bar{X} , H)
 \\& + (1-M) \odot log(1-D_{II}(\bar{X}, H)))_{ij} ]
 \\& + \alpha  [\sum_{i,j}((X \odot M)_{ij}log(\frac{(X \odot M)_{ij}}{(G(\tilde{X}, M, Z)V \odot M)_{ij}})
\\&-(X \odot M)_{ij}+(G(\tilde{X}, M, Z)V \odot M)_{ij})]
\\with \ \ & U_D = (YI) \odot U_p+(1-YI) \odot G(\tilde{X}, M, Z),
 \\&\bar{X}=\tilde{X}\odot M + MCL(G(\tilde{X}, M, Z)V) \odot (1-M) 
\end{split}
$}
\end{equation} 
Similar to a standard GAN, we alternate between training the generator and discriminators separately. The MF loss (\ref{loss1}) is only connected to the generator and will be utilized for multi-loss training alongside the generator's objective.

\section{Theoretical Discussion}  \label{theory}
\subsection{Global Optimality}
Without loss of generality, we assume the elements of $X$ as $X_{ij}=f(i,j)+s_{ij}$. Here, $f$ represents the bias effect of coordinates on X. The greater the internal correlation of $X$'s elements, the stronger regularity $f$ shows. $\{s_{ij}: i \in [0,m-1] \, j \in {0, n-1} \}$ are a series of random variables which follow independent and identical distribution $p_{d}$. We can define $\hat{X}$ in same way, i.e., 
$\hat{X}_{ij}=\hat{f}(i,j)+\hat{s}_{ij}$, where $\hat{s}$'s probability distribution is $p_g$. In real-world data, whether an element is masked or not may be affected by the value of the element itself. It also has different effects on the corresponding elements of $\hat{X}$ in generative model. So we need to discuss the probability distribution under different mask conditions. We assume that the distributions of $s$ corresponding to the unmasked elements in $X$ and $\hat{X}$ are $p_{d|m=1}$ and $p_{g|m=1}$, and to other masked elements are $p_{d|m=0}$ and $p_{g|m=0}$. The objective of this problem can be written as
\begin{equation}
\label{eq1} 
\hat{f} \to f \ \ and \ \  
p_{g|m=0}  \to p_{d|m=0} 
\end{equation} 
Throughout this section, we assume that the data are at MCAR (missing completely at random)\cite{heitjan1996distinguishing}, i.e. $M$ is independent of $X$. Then we can obtain $p_{d|m=0}=p_{d|m=1}=p_{d}$, and (\ref{eq1}) can be transformed into 
\begin{equation}
\label{eq7}
\hat{f} \to f \ \ and \ \
p_{g|m=0} \to p_{d|m=1},
\end{equation} 
or
\begin{equation}
\label{eq8}
\hat{f} \to f \ \ and \ \
p_{g|m=1} \to p_{d|m=1} \ \   and \ \  p_{g|m=0} \to p_{g|m=1}.
\end{equation} 
Obviously (\ref{eq1}), (\ref{eq7}) and (\ref{eq8}) are equivalent.

Below we will prove that the theoretical global optimal of (\ref{loss3}) is the same as that in (\ref{eq7}). The loss function for standard GAN is
\begin{equation} 
E_{x \sim p_{d}}[logD(x)]+ E_{z \sim p_g} [log(1-D(G(z)))] 
\end{equation} 
For this question, it can be re-written as:
\begin{equation} 
\label{eq4}
\begin{split}
&E_{\bar{S} \sim p_{d|m=1}}[logD(\bar{X} , H)]+ E_{\bar{S} \sim p_{g|m=0}} [log(1-D(\bar{X} , H))] 
\\& =  \int_{x}p_{d|m=1}(x-f)logD(x , H) 
\\&+ p_{g|m=0}(x-\hat{f})log(1-D(x , H))\mathrm{d}x
\end{split}
\end{equation} 
For any $a,b>0$, the function $alogx+blog(1-x)$ reaches its minimum if and only if $x=\frac{a}{a+b}$. So if G is fixed, the optimal discriminator D is $\frac{p_{d|m=1}(x-f)}{p_{d|m=1}(x-f)+p_{g|m=0}(x-\hat{f})}$. Substituting the optimal discriminator into (\ref{eq4}), we can obtain
\begin{equation} 
\label{eq41}
\resizebox{.97\linewidth}{!}{$
\begin{aligned}
&E_{\bar{S} \sim  p_{d|m=1}}[logD(\bar{X} , H)]
 + E_{\bar{S} \sim p_{g|m=0}} [log(1-D(\bar{X} , H))] 
\\ &= KL(p_{d|m=1}(x-f)
 \|\frac{1}{2}(p_{d|m=1}(x-f)+p_{g|m=0}(x-\hat{f})))
\\  & +KL(p_{g|m=0}(x-\hat{f}) 
\| \frac{1}{2}(p_{d|m=1}(x-f)+p_{g|m=0}(x-\hat{f})))
\\  & -log(4)
\end{aligned}
$}
\end{equation} 
We recognize that (\ref{eq4}) is essentially optimizing Kullback-Leibler divergence. It will achieve its global minimum $-log(4)$ if and only if $\hat{f}=f \  \& \  p_{g|m=0}=p_{d|m=1}$, which is the same as (\ref{eq7}). 

Next, we will calculate the theoretical global optimal of (\ref{loss1}). To do this, we initially introduce the entropy of the variable, $Ent(A)$, a measure of the disorder of $A$. According to the definitions of entropy and mutual information, we can find that
\begin{equation}
\begin{split}
Ent(\hat{X}) =& Ent(\hat{X} \odot M + \hat{X} \odot (1-M))
\\=& Ent(\hat{X} \odot M) + Ent(\hat{X} \odot (1-M))
\\&- I(\hat{X} \odot M, \hat{X} \odot (1-M))
\end{split}
\end{equation} 
where $I$ denotes mutual information. Noted that $I(\hat{X} \odot M , \hat{X} \odot (1-M)) \leq Ent(\hat{X} \odot (1-M))$ and the equality holds when $p_{g|m=0}=p_{g|m=1}$, which means that 
if $p_{g|m=1}$ is fixed, the global optimal of $min(Ent(\hat{X}))$ is $p_{g|m=0}=p_{g|m=1}$. Therefore, (\ref{eq8}) is equivalent to a multi-objective optimization problem
\begin{equation} 
\label{eq9}
min \  [D(X \odot M ||\hat{X} \odot M) , \   Ent(\hat{X})]
\end{equation} 
Where $D(\hat{X} \odot M||X \odot M)$ is the primary objective, which can be Euclidean distance or Kullback-Leibler divergence. Its theoretical optimal solution is $\hat{f}=f \ \& \  p_{g|m=1}=p_{d|m=1}$ under the assumption that $\hat{X}$ could be fitted by any functions (actually not). $Ent(\hat{X})$ is the second objective. We assume the global optimality of (\ref{eq9}) is $p_{g|m=0}=p_{g|m=1}=p_{d|m=1}=p^{*}$ , in which case $Ent(\hat{X}) = Ent^{*}(\hat{X})$. Thus, we have shown that problem (\ref{eq8}) is equivalent to the multi-objective optimization problem (\ref{eq9}).

Actually, it is very difficult to quantify $Ent(\hat{X})$ in (\ref{eq9}) in practice. Matrix factorization is essentially a constraint on entropy in the form of the network structure. From this premise, we target at solving (\ref{eq9}) approximately, and the optimization problem for the matrix factorization can be written as:
\begin{equation}
\begin{split}
\label{eq901}
min  \ \ &D(X \odot M ||\hat{X} \odot M)
\\s.t. \ \  &Ent(\hat{X}) < ent
\end{split}
\end{equation} 
Where $ent$ is a hyperparameter decided by the network structure (mainly $h$ mentioned in section \ref{mfa}). We proceed with the following two cases:
1. $ent>Ent^{*}(\hat{X})$. In this case, the optimal solution of  $\hat{f}=f \ \& \ p_{g|m=1} \to p_{d|m=1}$ can be reached theoretically, but it lacks a sufficient solution for $p_{g|m=0} \to p_{g|m=1}$. It corresponds to overfitting in general machine learning models. 2. $ent<Ent^{*}(\hat{X})$. In this case, $\hat{x}$ cannot be fully represented by the model due to its low degree of freedom limited by the tighter constraint on entropy, which hinders $\hat{x}$ from reaching the optimal of both $\hat{f}=f \ \& \ p_{g|m=1} \to p_{d|m=1}$ and $p_{g|m=0} \to p_{g|m=1}$. It corresponds to underfitting in general machine learning models. In either case, the theoretical global optimal of MF, i.e., problem (\ref{eq901}), is the approximate solution of (\ref{eq9}).

 \begin{table*}[h]
    \centering  
  
    \begin{tabular}{c|c|rrrrr}
    \toprule
    { } &\textbf{Method}&\textbf{ME-LA}&\textbf{PE-BAY}&\textbf{Cov-ca}&\textbf{Cov-de}&\textbf{Movie}\\
    \midrule

    { } &MissForest&0.2166&0.1037&0.1106&0.1589&-- \\
    \textbf{Block-wise}&Matrix Factorization&0.1637&0.0814&0.0645&0.0802&-- \\
    \textbf{Missing}&GAIN&0.2020&0.1135&0.0642&0.0811&--  \\
    \textbf{Data}&PC-GAIN&0.2169&0.1340& 0.1055 & 0.1166&-- \\
    { } &STGAN&0.2057&0.1229&0.0757&0.0968&-- \\
    { } &\textbf{BlockEcho(ours)}&\textbf{0.1561}&\textbf{0.0743}&\textbf{0.0495}&\textbf{0.0784}&--\\
    \midrule
    
    { } &MissForest&0.1328&0.0612&0.0493&0.0584&0.1998 \\
    \textbf{Scattered}&Matrix Factorization&0.1633&0.0816&0.0593&0.0737&0.1821 \\
    \textbf{Missing}&GAIN&0.1583&0.0664&0.0471&0.0616&0.2248 \\
    \textbf{Data}&PC-GAIN&0.1327&0.0617&0.0472&0.0673&0.1937 \\
    { } &STGAN&0.1444&0.0679&0.0351&\textbf{0.0470}&0.1983  \\
    { } &\textbf{BlockEcho(ours)}&\textbf{0.1322}&\textbf{0.0608}&\textbf{0.0321}&0.0488&\textbf{0.1797}\\
    \toprule
  \end{tabular}
  
  \caption{RMSE performance comparison of different data imputation methods at a fixed missing rate of 60\%. Noted that in Movie dataset, 80\% of the data are inherently missing, so we choose to mask the remaining visible elements at a missing rate of 60\%. 
  }
  \label{result_sg}
\end{table*}

\begin{figure*}[h]
  \centering
  \begin{subfigure}[b]{0.4\textwidth} 
    \includegraphics[width=\textwidth]{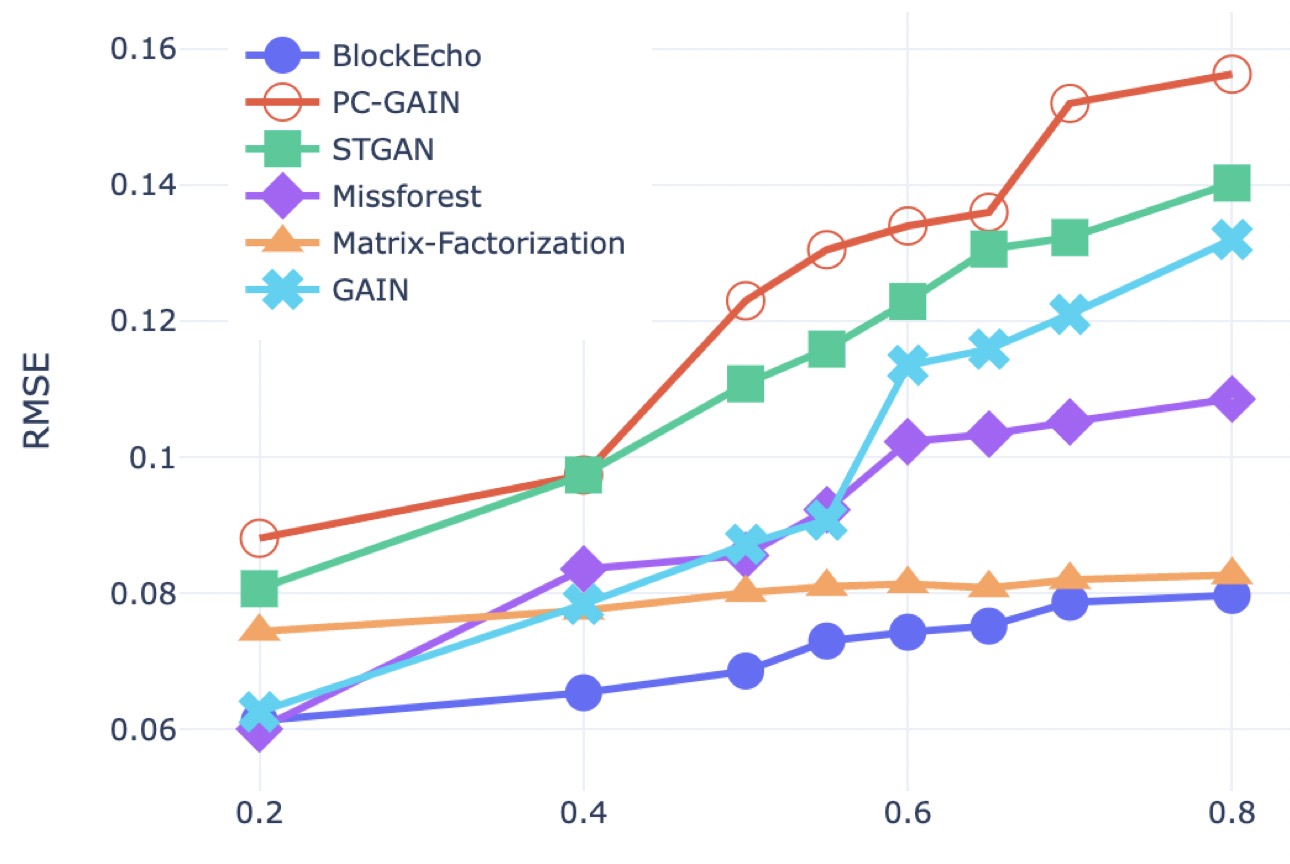} 
    \caption{Block-wise missing data} 
    \label{fig:subfig1} 
  \end{subfigure}
  \hspace{0\textwidth} 
  \begin{subfigure}[b]{0.405\textwidth}
    \includegraphics[width=\textwidth]{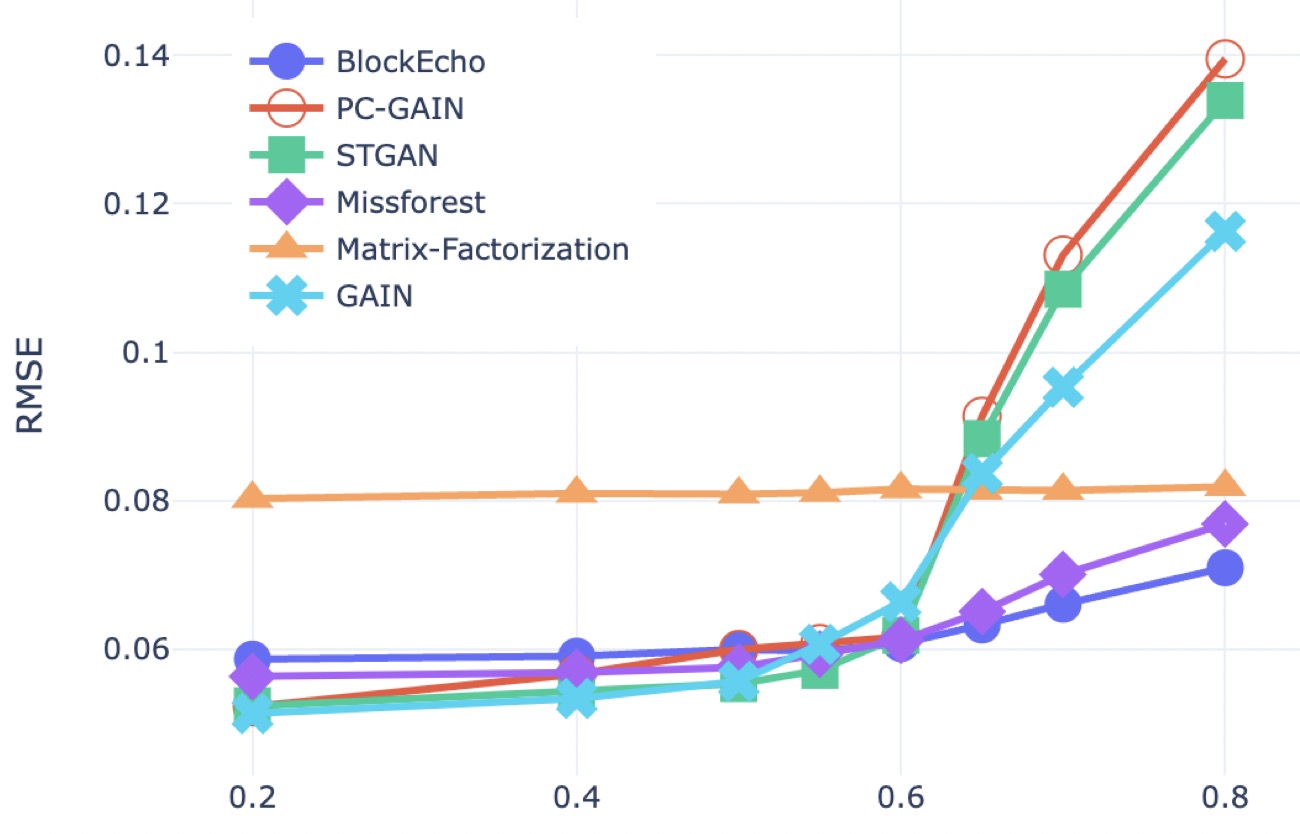}
    \caption{Scattered missing data}
    \label{fig:subfig2}
  \end{subfigure}
  \caption{RMSE performance trend with increasing missing rate} 
  \label{fig:table2}
\end{figure*}

\subsection{Convergence}
In the previous subsection, we demonstrated that loss (\ref{loss1}) has a similar but relatively inferior theoretical optimal solution compared to loss (\ref{loss3}), even when we introduce non-linear transformations to MF through neural networks. However, it is anticipated intuitively (also be verified by ablation experiments in subsection \ref{abl}), that the long-range dependencies introduced by MF will exert a significant influence on the outcomes of block-missing data imputation. This is due to the inherent challenge faced by models with a higher degree of freedom in effectively converging to fit long-range dependencies, particularly in the case of GAN models.
The convergence capacity of GANs has posed a substantial challenge since its inception. 
In 2017, L Mescheder et al. \cite{mescheder2017numerics} postulated that poor convergence is attributed to the eigenvalues of the Jacobian matrix of the gradient vector field having a zero real part and an excessively large imaginary part. Consequently, strategies such as gradient penalty \cite{gao2020data} and spectral normalization \cite{miyato2018spectral} have been adopted to alleviate this limitation. In tackling the specific matrix completion problem at hand, we leverage (\ref{loss1}), the MF loss function, and (\ref{loss2}), which enables the generator to more accurately emulate MF, to effectively steer the convergence of (\ref{loss3}).


At last of the section, we will prove the convergence of loss (\ref{loss1}), indicating that $D(\hat{X} \odot M||X \odot M)$ is nonincreasing under certain update rules.

 Let $F$ denote this expression. Let $u$ denote the set of parameters of the model, and $u^t$ the set of parameters upon update at step $t$. We introduce an auxiliary function $G(u, u^t)$ for $F(u)$ satisfying $G(u, u)=F(u)$ and $G(u, u^t) \geq F(u)$.Consequently, we can obtain $F(u^{t+1}) \leq F(u^{t})$ when $G(u^{t+1},u^t) \leq G(u^t,u^t)$, then $F$ is nonincreasing under the update  
 $u^{t+1} = arg min_{u} G(u,u^t)$.

 In our paper, we choose Kullback-Leibler divergence as the loss (\ref{loss1}) because the datasets are closer to Poisson distributions. In 2001, Daniel et al.\cite{lee2000algorithms} proposed the general auxiliary function for divergence. For this problem, we can make $F(u)$ equal to loss (\ref{loss1}) as $\hat{X}_{i,j}:=\sum_{a}u_{ia}v_{aj}$ and $\hat{X}_{i,j}^t:=\sum_{a}u_{ia}^tv_{aj}$, and define $G(u,u^t)$ as

 \begin{equation}
\begin{split}
&G(u,u^t) :=  \sum_{i,j}(((X \odot M)_{ij}log(X \odot M)_{ij}
\\&-(X \odot M)_{ij}+(\hat{X} \odot M)_{ij})
\\&-\sum_{a}(X \odot M)_{ij}\frac{u_{ia}^tv_{aj}}{\hat{X}_{i,j}^t}(logu_{ia}v_{aj}-log\frac{u_{ia}^tv_{aj}}{(\hat{X} \odot M)_{ij}^t}))
\end{split}
\end{equation} 

It is straightforward to verify that $G$ satisfies $G(u, u)=F(u)$ and $G(u, u^t)>=F(u)$ , and that $F(u^{t+1}) \leq G(u^{t+1},u^t) \leq G(u^t,u^t) \leq F(u^{t})$. Hence, $F$ is nonincreasing under certain update rules and the convergence of loss (\ref{loss1}) is proved. 



 \begin{table*}[h]
    \centering  
  
    \begin{tabular}{c|c|rrrrr}
    \toprule
    { } &\textbf{Method}&\textbf{ME-LA}&\textbf{PE-BAY}&\textbf{Cov-ca}&\textbf{Cov-de}&\textbf{Movie}\\ 
    \midrule

    { } &G+$D_I$+loss \ref{loss1}&0.1654&0.0793&0.0529&0.0798&-- \\
    \textbf{Block-wise} &G+$D_{II}$+loss \ref{loss1}&0.1708&0.0825&0.0548&0.0879&-- \\
    \textbf{Missing}  &G+$D_I$+$D_{II}$&0.2651&0.2082&0.1945&0.2261&-- \\
    \textbf{Data} &G+$D_I$+$D_{II}$+MSE loss&0.1734&0.0818&0.0513&0.0797&-- \\
    { } &\textbf{Ours}(G+$D_I$+$D_{II}$+loss \ref{loss1})&\textbf{0.1561}&\textbf{0.0743}&\textbf{0.0495}&\textbf{0.0784}&-- \\
    \midrule
    
    { } &G+$D_I$+loss \ref{loss1}&0.1409&0.0631&0.0346&0.0510&0.1916 \\
    \textbf{Scattered} &G+$D_{II}$+loss \ref{loss1}&0.1419&0.0693&0.0408&0.0585&0.1921 \\
    \textbf{Missing}  &G+$D_I$+$D_{II}$&0.2071&0.1602&0.1322&0.1766&0.2810  \\
    \textbf{Data} &G+$D_I$+$D_{II}$+MSE loss&0.1398&0.0668&0.0363&0.0564&0.1893 \\
    { } &\textbf{Ours}(G+$D_I$+$D_{II}$+loss \ref{loss1})&\textbf{0.1322}&\textbf{0.0608}&\textbf{0.0321}&\textbf{0.0488}&\textbf{0.1797} \\
    \bottomrule
  \end{tabular}
  
  \caption{RMSE performance in the ablation study. The design of experiments is the same as \ref{p_p}
  }
  \label{result_as}
\end{table*}

\section{Experiments and Evaluation} \label{experiment}

In this section, we rigorously test the performance of our data imputation method, BlockEcho, through a series of four carefully designed experiments using various real-world datasets available in the public domain. The initial set of experiments compare BlockEcho with state-of-the-art baselines, focusing on a fixed missing rate as high as 60\%. Following this, further experiments are conducted to assess the stability of these models as the missing rate is incrementally increased. Additionally, an ablation study is performed to examine the impact of specific components of the method. Finally, we designed a downstream prediction task using the imputed data as input, to evaluate how the performance of the data imputation method influences the end-to-end performance of a real-world prediction task.
The experiments in this project are based on the following open-source real-world datasets from three distinct fields:

\begin{enumerate}
\item \emph{Traffic Datasets (Time-Street matrix)}. We employ two traffic flow datasets, ME-LA and PE-BAY, 
which collected 
in the highway of Los Angeles County and from California Transportation Agencies Performance Measurement System\cite{li2017diffusion}.
The traffic data are characterised by strong periodicity such as peak hours.
\item \emph{COVID-19 Dataset (Date-City matrix)}. This government dataset records daily COVID-19 cases (Cov-ca) and deaths (Cov-de) in major cities around the world since 2020. Infectious disease datasets are characterized by local continuity and sudden bursts. 
\item \emph{Movie Dataset (User-Movie matrix)}. This dataset (Movie) records movie ratings collected from the MovieLens website\cite{harper2015movielens}, which contains a large amount of missing data.
\end{enumerate}

For ``scattered missing data," we generate a random matrix of the same size as the original data and rank the random numbers within it. We compare the rankings with $miss rate*m*n$ to determine which original elements should be masked. As for the ``block-wise" missing data, we randomly select the top-left and bottom-right corners of a ``block" within the feasible region limited by the missing rate, and then mask the original elements within that region.

All models were trained on an Nvidia Tesla V100S PCIE GPU and each experiment is repeated for ten times with different random seeds, and the results are averaged. 

 \subsection{Performance of BlockEcho} \label{p_p}
In the first series of experiments, we synthetically mask 60\% of the data in each dataset to generate a missing data matrix, and we compare the performance of BlockEcho with 5 baseline matrix completion models -  
MissForest\cite{2011MissForest}, Matrix Factorization\cite{hastie2015matrix}, GAIN\cite{yoon2018gain}, PC-GAIN \cite{wang2021pc} and STGAN\cite{yuan2022stgan}
— in terms of imputation accuracy. Data are masked in two ways: block-wise mask and scattered mask. To measure the imputation accuracy, we use RMSE to calculate the error between the model-imputed data and the original masked data, which is defined as
$RMSE = \frac{||(\bar{X}-X) \odot (1-M)||_{F}}{\sum_{i,j}(1-M)_{ij}} $
. 
In order to have a relatively uniform measure for different datasets, we normalize the data before computing RMSE. Table \ref{result_sg} 
reports the RMSE for BlockEcho and 5 other imputation models. Results show that BlockEcho performs significantly better than baselines, especially for block-wise missing data.

\subsection{Models performance in different missing rates} 


In the second series of experiments, we take the traffic dataset PE-BAY as a representative, dynamically adjust the data missing rate from 20\% to 80\%, and explore the variation trend of each model performance with the data missing rate.  Figure  \ref{fig:table2} quantitatively shows this trend: the SOTA machine learning algorithms represented by GAIN perform better on datasets with a low data missing rate, but will deteriorate rapidly as the missing rate increases; the Matrix Factorization method performs more stable but has a lower ceiling for accuracy. BlockEcho absorbs the advantages of both, and thus gives the best performance at high missing rates. Although at low missing rates it is not as accurate as some SOTA models, BlockEcho significantly outperforms them at high missing rates when the datasets become more incomplete. This superiority is more obvious in block-wise missing datasets.

\subsection{Ablation study} \label{abl}
In this subsection we design ablation experiments to verify the contribution of each component of BlockEcho to the results. The table \ref{result_as} shows that when we remove any part or replace it with a more conventional alternative, it will negatively affect the results. It is worth mentioning that when the matrix factorization loss function \ref{loss1} is removed and only the GAN framework is used for matrix completion, the accuracy of the imputed data will be significantly reduced, which is why most generative models (including ours) use direct loss functions to guide the convergence of GAN.

\subsection{Case study: traffic forecasting} 
In the final series of experiments, we design a downstream prediction task to illustrate how the performance of the data imputation method affects the end-to-end performance of a real-world prediction task. 
We take PE-BAY dataset as input to forecast the traffic conditions at the next timestamp. 
For the input datasets, we take the original data, the imputed data with our model BlockEcho, and the imputed data from each baseline model for comparison.
For traffic forecasting, we use Random Forests (RF) with the same set of hyperparameter settings with the same feature engineering for all input datasets. 
We use Weighted Mean Absolute Percentage Error (WMAPE) as the error metric to measure the prediction performance. 
Table \ref{result_pm} summarizes the prediction results. Comparing Table \ref{result_sg}  and Table \ref{result_pm}, we find that models with higher data imputation accuracy tend to give better results in subsequent prediction tasks, and BlockEcho, the best-performing data imputation model, also gives the lowest prediction errors in subsequent traffic forecasting tasks.

 \begin{table}[htbp]
    \centering  
    \begin{tabular}{l|rr}
    \toprule
    {} &Block-wise&Scattered \\ 
    \midrule
    
    \textbf{Ori-Data}&\textbf{0.0378}&\textbf{0.0378} \\
    MissForest&0.0696&0.0512 \\
    Matrix Factorization&0.0615&0.0619 \\
    GAIN&0.0789&0.0484 \\
    PC-GAIN&0.0702&0.0479\\
    STGAN&0.0685&0.0503 \\
    \textbf{BlockEcho(ours)}&\textbf{0.0484}&\textbf{0.0473}\\
    \bottomrule
  \end{tabular}
  
  \caption{WMAPE of prediction task after various imputation model. 
  }
  \label{result_pm}
\end{table}

\section{Conclusion and Future Work} \label{conclusion}
In this paper, we excavate and mathematically define the issue of ``block-wise” missing data and innovatively propose the solution, BlockEcho. Experiments on various data sets, especially with block missing data, show that our method outperforms other SOTA methods. Our future work will extend to federated learning where block-wise missing data widely appear.

\appendix

\bibliographystyle{named}
\bibliography{ijcai24}

\end{document}